\pdfoutput=1
\documentclass[11pt]{article}

\usepackage[]{acl}

\usepackage{times}
\usepackage{latexsym}

\usepackage[T1]{fontenc}
\usepackage[utf8]{inputenc}

\usepackage{microtype}

\usepackage{multirow}
\usepackage{amsmath}
\usepackage{amsfonts}
\usepackage{amssymb}
\usepackage{hyperref}
\hypersetup{
    colorlinks=true,
    linkcolor=blue
}

\usepackage{booktabs}

\usepackage{paralist}

\usepackage{mdwlist}
\usepackage{graphicx}

\usepackage{color}
\definecolor{red}{RGB}{200,0,0}
\definecolor{green}{RGB}{0,100,0}
\definecolor{blue}{RGB}{0,0,200}
\usepackage{bbm}
\usepackage{soul}

\usepackage[ruled,linesnumbered]{algorithm2e}

\usepackage{xspace}

\newcommand{\method}{RePP\xspace}
\newcommand{\baseline}{Vanilla\xspace}

\title{Zero-shot Domain Adaptation for Neural Machine Translation with Retrieved Phrase-level Prompts}

\author{
Zewei Sun\textsuperscript{\rm 1}, 
Qingnan Jiang\textsuperscript{\rm 1}, 
Shujian Huang\textsuperscript{\rm 2},
Jun Cao\textsuperscript{\rm 1}, 
Shanbo Cheng\textsuperscript{\rm 1}, 
Mingxuan Wang\textsuperscript{\rm 1}\\
\textsuperscript{\rm 1} ByteDance AI Lab, 
\textsuperscript{\rm 2} State Key Laboratory for Novel Software Technology, Nanjing University \\
\texttt{\{sunzewei.v,jiangqingnan.k,caojun.sh\}@bytedance.com} \\
\texttt{\{chengshanbo,wangmingxuan.89\}@bytedance.com,}
\texttt{huangsj@nju.edu.cn}
}

\begin{document}
\maketitle
\begin{abstract}
% Prompt-based learning has been a new paradigm for adapting pre-trained models.
% % In this paradigm, pre-trained models learn from in-context examples by constructing prompts for specific task prediction.
% Two main difficulties hinder its application in machine translation. First, it is hard to construct effective prompts for machine translation. Second, the task inconsistency of pre-training and prompt-based prediction limits the potential of this paradigm.
% To tackle the two difficulties, we propose bilingual phrase prompts and prompt-aware pre-training for prompt-based machine translation.
% Experiments show that the proposed methods improve domain-specific machine translation for 6.2 BLEU scores and lexically constrained machine translation for 11.5\% accuracy without additional training.

Domain adaptation is an important challenge for neural machine translation. However, the traditional fine-tuning solution requires multiple extra training and yields a high cost. In this paper, we propose a non-tuning paradigm, resolving domain adaptation with a prompt-based method. Specifically, we construct a bilingual phrase-level database and retrieve relevant pairs from it as a prompt for the input sentences. By utilizing \textbf{Re}trieved \textbf{P}hrase-level \textbf{P}rompts (\textbf{\method}), we effectively boost the translation quality. 
Experiments show that our method improves domain-specific machine translation for 6.2 BLEU scores and improves translation constraints for 11.5\% accuracy without additional training.

\end{abstract}

\section{Introduction}
\label{sec:intro}
In the past years, neural machine translation (NMT) has shown its great power~\cite{bahdanau2015neural,vaswani2017attention}. However, domain adaptation is still a challenge for NMT~\cite{koehn2017six}. The traditional solution is fine-tuning on domain-specific data. However, this requires multiple extra training. Every time we encounter a new domain, the tuning procedure needs to be conducted again. In this paper, we propose to resolve domain adaptation with a prompt-based method in a zero-shot style.

Prompt-based learning has been an attractive method for adapting pre-trained models to specific tasks in recent years. With handcrafted or automatically created prompts, pre-trained models can achieve good performance in many downstream tasks without fine-tuning~\cite{schick2020few,schick2021exploiting,schick2020automatically}.

% Pre-trained models (PTMs) have promoted natural language processing (NLP) significantly~\cite{devlin2019bert,radford2018improving,radford2019language,brown2020language}.
% In recent years, prompt-based learning has been an attractive method for adapting PTMs to specific tasks.
% With handcrafted or automatically created prompts, PTMs could achieve good performance in many downstream tasks without fine-tuning~\cite{schick2020few,schick2021exploiting,schick2020automatically}.

Among them, machine translation can also be potential~\cite{brown2020language,garcia2022using}. With appropriate prompts, we can improve the translation results without traditional tuning in some data-scarce scenes such as domain adaptation. However, the construction of prompts can be difficult.~\citet{liu2021what,sun2022rethinking} reveal that the performance of downstream tasks relies on the selection of in-context examples heavily. But it is hard to find relevant translation examples since sentences are very sparse.
% \citet{brown2020language} proposed to construct the prompt by concatenating some in-context translation examples. 
% \citet{liu2021what} revealed that the performance of downstream tasks relies on the selection of in-context examples heavily.
% Second, the prompt-based training are not co-designed with normal machine translation.
% The inconsistency between pre-training and prediction limits the potential of this paradigm.

% \begin{figure}
%     \centering
%     \includegraphics[width=7.5cm]{fig/motivation.pdf}
%     \caption{motivation.}
%     \label{fig:papt}
% \end{figure}

% \begin{figure}
%     \centering
%     \includegraphics[width=7.5cm]{fig/papt.pdf}
%     \caption{The illustration of vanilla training and prompt-aware co-training.}
%     \label{fig:papt}
% \end{figure}

% \begin{table}[]
% \small
%     \centering
%     \begin{tabular}{c}
%     \hline
%          \texttt{<source> -> <target>} \\
%     \hline
%          \texttt{<source> -> <target>} \\
%          \texttt{<prompt> <p> <source> -> <target>} \\
%     \hline
%     \end{tabular}
%     \caption{The illustration of prompt aware pre-training. For each translation example, we create a new training example by concatenating the prompt with the source sentence. \texttt{<p>} is a special token that separates the prompt and the source sentence.}
%     \label{tab:prompt-aware-pretraining}
% \end{table}

\begin{figure*}
    \centering
    \includegraphics[width=14cm]{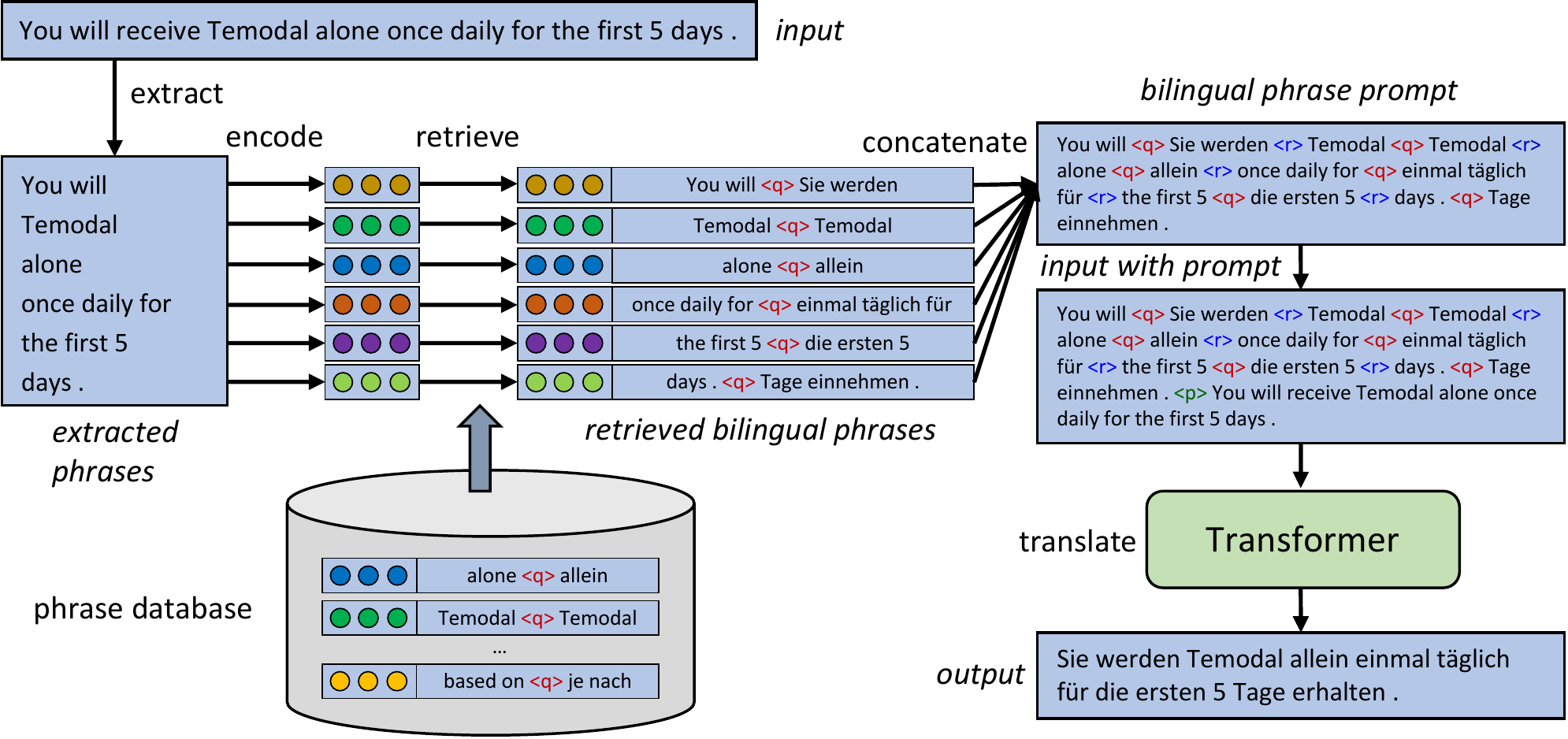}
    \caption{Overview of constructing bilingual phrase prompts for prompt-based machine translation prediction.}
    \label{fig:overview}
\end{figure*}

% \begin{table*}[]
% \small
%     \centering
%     \begin{tabular}{l}
%     \hline
%          Source: \texttt{You will receive Temodal alone once daily for the first 5 days .} \\
%     \hline
%         Extract phrases from source: \\ 
%         \texttt{You will}; \quad \texttt{Temodal}; \quad \texttt{alone}; \quad \texttt{once daily for}; \quad \texttt{the first 5}; \quad \texttt{days .} \\
%     \hline
%         Retrieve relevant bilingual phrases: \\
%         \texttt{You will} $\rightarrow$ \texttt{You will <q> Sie werden}  \\
%         \texttt{alone} $\rightarrow$ \texttt{alone <q> allein} \\
%         \texttt{once daily for} $\rightarrow$ \texttt{once daily for <q> einmal täglich für} \\
%         \texttt{the first 5} $\rightarrow$ \texttt{the first 5 <q> die ersten 5} \\
%         \texttt{days .} $\rightarrow$ \texttt{days . <q> Tage einnehmen .} \\
%     \hline
%         Construct prompt with relevant bilingual phrases: \texttt{You will <q> Sie werden <r> Temodal <q> Temodal} \\ \texttt{<r> alone <q> allein <r> once daily for <q> einmal täglich für <r> the first 5 } \\
%         \texttt{<q> die ersten 5 <r> days . <q> Tage einnehmen .} \\
%     \hline
%      \end{tabular}
%     \caption{The process of retrieving source-relevant bilingual phrases to construct a prompt.}
%     \label{tab:process}
% \end{table*}

In this work, we extract phrase-level bilingual prompts to tackle this problem since phrases are much denser than sentences. We build a bilingual phrase-level database which contains phrase pairs and corresponding contextual representation. In the inference stage, we retrieve input-relevant phrases from the database as a prompt. By appending the prompt to the source sentence, we try to improve the translation results.
% And we propose to obtain input-relevant prompts with retrieve-based methods.
% By retrieving bilingual phrases from a pre-built bilingual phrase database, we can construct corresponding prompts, which provides useful knowledge for the translation generation. 

% We also use some data augmentation techniques in the training phase to make the model acquire the ability to handle prompt-aware inputs. For each example, we create a new training example by appending a prompt to the input sentence. 
% Then we train a sequence-to-sequence model on both original data and new constructed data. As a results, the trained models could translate sentences with or without prompt.

In the end, we can influence the translation model positively and quickly. On the one hand, relevant prompts can bring in correct translation, especially in unseen domains. On the other hand, the prompt-based methods need no fine-tuning procedure, making the improvements fast and low-cost. Finally, our method can guarantee the data privacy for users since no corpus is exposed to the model except for filtered phrase pairs.

Experimental results show that our method outperforms the baseline for 6.2 BLEU scores on average in domain-specific machine translation and gains 11.5\% accuracy boost in lexical constrained machine translation. It is worth noting that these improvements all come without any additional training.

% In this work, we propose effective methods to tackle the two difficulties.
% First, we propose to construct bilingual phrase prompts for machine translation. 
% The prompts is concatenated by phrase-level translation examples, i.e. bilingual phrases, to alleviate the sparsity of sentence-level translation examples.
% By retrieving relevant bilingual phrases from a pre-built bilingual phrase database, we construct input-relevant prompt, which provides useful knowledge for translation generation.
% Then, to alleviate the inconsistency of pre-training and prompt-based prediction, we propose to pre-train prompt-aware models for machine translation.
% As Figure \ref{fig:papt} shows, for each example, we create a new training example by prefixing a prompt to the input sentence. 
% Then we pre-train encoder-decoder models on both original training data and new created data.
% The trained models could translate sentences with or without prompt.

% Our contributions are summarized as follows:

% \begin{itemize}
%     \item We propose bilingual phrase prompts and prompt-aware pre-training to tackle difficulties of applying prompt-based learning in machine translation.
%     \item The proposed method outperforms the baseline for 6.2 BLEU scores in domain-specific machine translation and 11.5\% accuracy in lexical constrained machine translation on average without additional training.
% \end{itemize}

\section{Related Work}
\label{sec:related}
Prompt-based learning is a new paradigm of adapting pre-trained language models to specific tasks.
Different from fine-tuning and feature-based adaptation~\cite{devlin2019bert,peters2018deep}, prompt-based learning does not need additional training on downstream tasks.
It formulates downstream tasks to language model slot filling tasks~\cite{liu2021pre} with prompts.
Generally, in prompt-based learning, making predictions on specific tasks with pre-trained language models contains 3 stages. 
(i) construct a prompt based on the input, which has some unfilled slots;
(ii) fill the unfilled slots with pre-trained models;
(iii) derive the final prediction from the filled slots.

The prompt formats depend on the pre-trained models and downstream tasks.
There are two main categories of prompts~\cite{liu2021pre}: cloze prompts where unfilled slots are pre-defined blanks, and prefix prompts where filling slots is continuing generation with the prefix.
Cloze prompts are usually used in natural language understanding tasks~\cite{schick2020few,schick2021exploiting,schick2020automatically,cui2021template} while prefix prompts are mainly used in natural language generation tasks~\cite{brown2020language}.

Our paper is also related to domain adaptation~\cite{chu2018survey,hu2019domain,bapna2019simple}, lexical constraint utilization~\cite{dinu2019training,chen2021lexical}, pre-trained models for NMT~\cite{sun2021multilingual}, and memory-based methods~\cite{cao2018encoding,he2021fast,cai2021neural}. Differently, we propose to use zero-shot prompt-based methods to tackle domain adaptation for neural machine translation.

\section{Methodology}
\label{sec:approach}

In this section, we first introduce how to construct the bilingual phrase database and how to retrieve the input-relevant prompts.
Then we illustrate the prompt-aware co-training which makes the model acquire the ability to translate sentences with or without prompts.
The overview of our method is shown in Figure \ref{fig:overview}.

% \begin{table*}[]
% \small
%     \centering
%     \begin{tabular}{c|ccccc|ccccc}
%     \hline
%     & \multicolumn{5}{c|}{\textbf{English $\rightarrow$ German}} & \multicolumn{5}{c}{\textbf{German $\rightarrow$ English}} \\
%          & \textbf{WMT14} & \textbf{Law} & \textbf{Medical} & \textbf{Koran} & \textbf{IT} & \textbf{WMT14} & \textbf{Law} & \textbf{Medical} & \textbf{Koran} & \textbf{IT} \\
%          Database size & 131.7M & 23.0M & 7.1M & 0.5M & 3.9M & 129.2M & 21.9M & 7.0M & 0.5M & 3.9M \\
%         \hline
%         \baseline & \textbf{27.43} & 33.60 & 29.95 & 10.70 & 24.37 & \textbf{32.24} & 39.07 & 34.24 & 12.20 & 31.14 \\
%         \method (w/o prompt) & 27.18 & 33.95 & 29.95 & 10.49 & 23.94 & 32.00 & 39.02 & 33.64 & 11.62 & 30.19  \\
%         \method (w/ prompt) & 25.94 & \textbf{42.37} & \textbf{40.81} & \textbf{14.21} & \textbf{28.09} & 29.98 & \textbf{45.87} & \textbf{44.18} & \textbf{14.32} & \textbf{32.44} \\
%     \hline
%     \end{tabular}
%     \caption{Test set BLEU scores in domain adaptation for machine translation. In general domain translation \method performs sightly worse than \baseline. In domain-specific translation, \method outperforms \baseline for 6.7 and 5.6 BLEU scores on average in English $\rightarrow$ German and German $\rightarrow$ English directions without additional training.}
%     \label{tab:main-results}
% \end{table*}

\begin{table*}[h]\footnotesize
% \small
    \centering
    \begin{tabular}{c|cccc|cccc}
    \hline
    Translation direction & \multicolumn{4}{c|}{\textbf{English $\rightarrow$ German}} & \multicolumn{4}{c}{\textbf{German $\rightarrow$ English}} \\
        Domain & \textbf{Law} & \textbf{Medical} & \textbf{Koran} & \textbf{IT} & \textbf{Law} & \textbf{Medical} & \textbf{Koran} & \textbf{IT} \\
         Database size & 23.0M & 7.1M & 0.5M & 3.9M & 21.9M & 7.0M & 0.5M & 3.9M \\
        \hline
        \baseline &  33.60 & 29.95 & 10.70 & 24.37 &  39.07 & 34.24 & 12.20 & 31.14 \\
        % \method (w/o prompt) &  33.95 & 29.95 & 10.49 & 23.94 & 39.02 & 33.64 & 11.62 & 30.19  \\
        \method (Zero-shot) & \textbf{42.37} & \textbf{40.81} & \textbf{14.21} & \textbf{28.09} & \textbf{45.87} & \textbf{44.18} & \textbf{14.32} & \textbf{32.44} \\ \hline \hline
        % Fine-tuning & 49.07 & 47.10 & 25.98 & 36.28 & 55.19 & 51.35 & 22.87 & 41.88\\
        Fine-tuning (Fully Supervised) & 45.02 & 44.52 & 15.43 & 34.48 & 50.95 & 47.48 & 18.13 & 39.57 \\
    \hline
    \end{tabular}
    \caption{Test set BLEU scores in domain adaptation for machine translation. Database size represents the number of bilingual phrases in the database. In domain-specific translation, \method outperforms \baseline for 6.7 and 5.6 BLEU scores on average in English $\rightarrow$ German and German $\rightarrow$ English directions without additional training.}
    \label{tab:main-results}
\end{table*}

\subsection{Construction and Retrieval of Phrase-level Prompts}

\paragraph{Bilingual phrase prompts:}
A bilingual phrase prompt is concatenated by a few bilingual phrases separated by \textcolor{blue}{<r>}.
Each bilingual phrase pair contains a source language phrase and the corresponding target language phrase translation separated by \textcolor{red}{<q>}, such as ``based on \textcolor{red}{<q>} je nach''.

We hypothesize that input-relevant bilingual phrases provide useful knowledge for machine translation.
For an input sentence, we retrieve the input-relevant bilingual phrases from a pre-built phrase database based on representation to construct a prompt.

\paragraph{Offline bilingual phrase database construction:}
We extract bilingual phrases from parallel translation data and compute the contextualized representations of the source language phrases with multilingual BERT~\cite{devlin2019bert}.
The contextualized representation and the corresponding bilingual phrases constitutes a key-value pair.
The phrase database is the dictionary of key-value pairs created from parallel translation data.
It is worth noting that there may exist different phrase pairs that share the same source phrase due to the ambiguity~\cite{sun2020generating}. They are distinguished by the contextual representation.

We extract bilingual phrases by first extracting word alignments with awesome-align~\cite{dou2021word} and then extracting bilingual phrases from word alignments with the algorithm described in \citet{koehn2010statistical}.
The contextualized representation of a phrase is computed by average-pooling on the hidden states of words in the phrase.

\paragraph{Online input-relevant prompts retrieval:}

Figure \ref{fig:overview} illustrates the process of constructing a prompt for an input sentence.
We extract phrases in the input sentence and compute the contextualized representations of these phrases with multilingual BERT. Using the representation as the searching key, we retrieve the most similar bilingual phrases from the database based on the $L^2$ distance.
% that exists in the database. However, due to the phrase ambiguity, one source phrase can exist in many source-target phrase pairs with different contexts in the database. Therefore, we compute the contextualized representations of phrases in the input sentence with multilingual BERT and retrieve the most similar bilingual phrases from the database based on the $L^2$ distance of the representation.
At last, we concatenate the retrieved bilingual phrases to construct a bilingual phrase prompt.

% We load the source phrases of phrase database into a trie and extract phrases in the input sentence that exists in the trie.

\subsection{Prompt-aware Training and Inference}

% \paragraph{Model architecture}

% We train an encode-decoder model for prompt-based learning in machine translation.
% % The encoder encodes a source sentence into a sequence of hidden states while the decoder generates a target sentence with the output hidden states.
% The encoder-decoder model employ the Transformer architecture~\cite{vaswani2017attention}.

\paragraph{Mixed training data:}

% \sout{The training data is created based on the general domain translation data.
% We build a phrase database from general domain translation data from input-relevant prompt mining.
% As Figure \ref{fig:papt} shows, for an input-output pair, we construct an input-relevant prompt and put it before the input sentence to create a new training example.
% To prevent retrieving bilingual phrases extracted from the current input-output pair and overfitting retrieved bilingual phrases,
% we construct prompt for training data by retrieving the second similar bilingual phrases.
% We pre-train the encoder-decoder model on both original training data and created training data with cross entropy loss.}

\begin{figure}[hb]
    \centering
    \includegraphics[width=7.5cm]{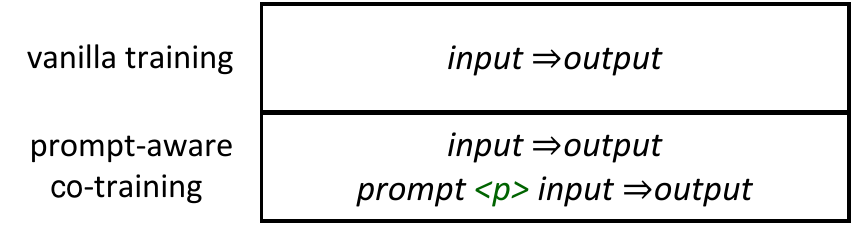}
    \caption{The illustration of vanilla training and prompt-aware co-training.}
    \label{fig:papt}
\end{figure}

To keep the consistency between prompt-aware inference and translation training, it is crucial to involve prompt patterns in the training phase. But we also need to maintain the normal translation ability in the model. Therefore, we mix the original data (general domain) with handcrafted prompt-aware data in the training, as is shown in Figure~\ref{fig:papt}. The idea of constructing prompt-aware data is similar to the inference. We retrieve some phrases in the general domain as a prompt and prepend it to the input sentence to yield a new training instance. 
% To prevent overfitting, the most similar phrase pair is omitted. 
It is worth noting that no in-domain data is involved so it is still zero-shot for domains.

\paragraph{Prompt-based prediction}

After co-training on the mixed data, the model can translate input sentences with or without prompts.
Also, the prompts can be created either with automatic input-relevant prompt mining or handcrafting for translation intervention purpose.
For example, we can provide bilingual phrases to manually construct a prompt to make a lexical constraint.
% For example, soft lexical constraints can be represented as bilingual phrase prompts to interfere the lexical selection during translation process.

% \begin{table}[]
%     % \small
%     \centering
%     \begin{tabular}{c|cccc}
%     \hline
%         Domain & Law & Medical & Koran & IT \\
%         \hline
%         Train & 467k & 248k & 18k & 223k \\
%         Dev & 2k & 2k & 2k & 2k \\
%         Test & 2k & 2k & 2k & 2k \\
%         \hline
%     \end{tabular}
%     \caption{The statistics of datasets used in our experiments.}
%     \label{tab:data-statistics}
% \end{table}

\section{Experiments}
\label{sec:experiments}

\subsection{Experimental Setup}

We evaluate the proposed method in English $\rightarrow$ German and German $\rightarrow$ English translation directions.
We train the models in general domain translation datasets.
Then we evaluate the domain-specific translation performance without additional tuning.
For prompt-based translation prediction, we mine input-relevant prompts automatically with in-domain phrase databases.

\subsection{Baseline}
We compare the proposed \method with baseline --- \baseline, which is a vanilla Transformer~\cite{vaswani2017attention} directly trained on the general domain datasets. The result of fully supervised fine-tuning is also listed for reference only.

\subsection{Datasets}

We take WMT14 EN-DE dataset as the general domain dataset for baseline and \method training.
We evaluate the domain-specific translation performance on 4 datasets\footnote{https://github.com/roeeaharoni/unsupervised-domain-clusters} proposed by \citet{koehn2017six} and re-splitted by \citet{aharoni2020unsupervised}. The training data in the specific domains is for prompt mining.
% The statistics of these datasets are shown in Table \ref{tab:data-statistics}.

\begin{table*}[]
    \small
    \centering
    \begin{tabular}{l}
    \hline
        \textbf{Source:} What is the risk associated with Poulvac FluFend H5N3 RG ? \\
        \textbf{Reference:} Welche Risiken sind mit Poulvac FluFend H5N3 RG verbunden ? \\
    \hline
        \textbf{No Prompt} \\
        \textbf{Translation:} Was ist das Risiko für Poulvac Fluf H5N3 RG ? \\
    \hline
        \textbf{Prompt:} What is the risk \textcolor{red}{<q>} \sethlcolor{pink}\hl{Welches Risiko ist} \\
        \textbf{Translation:} \sethlcolor{pink}\hl{Welches Risiko ist} mit Poulvac Fend H5N3 RG verbunden ? \\
    \hline
        \textbf{Prompt:} associated with \textcolor{red}{<q>} \sethlcolor{cyan}\hl{mit Pirsue verbunden} \\
        \textbf{Translation:} Was ist das \sethlcolor{cyan}\hl{mit} Poulvac Fend H5N3 RG \sethlcolor{cyan}\hl{verbundene} Risiko ? \\
    \hline
        \textbf{Prompt:} Poulvac FluFend H5N3 \textcolor{red}{<q>} \hl{Poulvac FluFend H5N3} \\
        \textbf{Translation:} Was ist das mit \hl{Poulvac FluFend H5N3} RG verbundene Risiko ? \\
    \hline
        \textbf{Prompt:} What is the risk \textcolor{red}{<q>} \sethlcolor{pink}\hl{Welches Risiko ist} \textcolor{blue}{<r>} associated with \textcolor{red}{<q>} \sethlcolor{cyan}\hl{mit Pirsue verbunden} \textcolor{blue}{<r>}  Poulvac FluFend \\ H5N3 \textcolor{red}{<q>} \hl{Poulvac FluFend H5N3} \textcolor{blue}{<r>} RG \textcolor{red}{<q>} \hl{RG} \textcolor{blue}{<r>} ? \textcolor{red}{<q>} \sethlcolor{yellow}\hl{?} \\
        \textbf{Translation:} \sethlcolor{pink}\hl{Welches Risiko ist} \sethlcolor{cyan}\hl{mit} \sethlcolor{yellow}\hl{Poulvac FluFend H5N3} \hl{RG} \sethlcolor{cyan}\hl{verbunden} \sethlcolor{yellow}\hl{?} \\
    \hline
    \end{tabular}
    \caption{An English$\rightarrow$German translation case with different bilingual phrase prompts. Bilingual phrase prompts enforce soft lexical constraints on the translation process.}
    \label{tab:case-study}
\end{table*}

\subsection{Implementation Details}

We use joint Byte Pair Encoding~\cite{sennrich2016neural} with 32k merge operations for subword segmentation.
We employ the Transformer Base architecture~\cite{vaswani2017attention,sun2020alleviating} for \method and baseline training.
We train all the models for ten epochs.
The models are implemented based on Fairseq~\cite{ott2019fairseq}\footnote{https://github.com/pytorch/fairseq}.
For efficient bilingual phrase retrieval, we build an IVFPQ index with FAISS~\cite{JDH17}\footnote{https://github.com/facebookresearch/faiss}.%, where each key vector is quantizated into 32 bytes.

\subsection{Results: Significant Improvements}

The results of our experiments are shown in Table \ref{tab:main-results}.
\method outperforms \baseline for 6.7 and 5.6 BLEU scores in English $\rightarrow$ German and German $\rightarrow$ English translation without any additional training.
The improvements indicate that incorporating in-domain bilingual phrase prompts can significantly help domain-specific machine translation.

Though \method does not perform as well as fine-tuning, we want to highlight that it needs no extra training, making it a fast and low-cost way to adapt to unseen domains. Besides, \method only needs one main model while fine-tuning has to maintain $n$ models for $n$ domains.

\section{Analysis}
\label{sec:analysis}

\subsection{Lexically Constrained Translation}

% Lexical constraints in machine translation can naturally represented as bilingual phrase prompts.
% We conduct analysis on the zero-shot lexically constrained translation ability of the proposed method.

% Lexically constrained machine translation has an additional constraint that a specific phrase \textbf{x} in the input sentence is translated to a target phrase \textbf{y} in the translation output.
% We represent this lexical constraint as a bilingual phrase prompt "\textbf{x} \textcolor{red}{<q>} \textbf{y}" and translate the input sentence with the prompt.

Lexical constraints in machine translation can be naturally represented as bilingual phrase prompts. In other words, a specific phrase \textbf{x} in the input sentence should be translated to a target phrase \textbf{y} in the translation output.

We test the lexically constrained translation ability of \baseline and \method in English $\rightarrow$ German translation.
We use the Wiktionary and IATE test sets created by \citet{susanto2020lexically} for evaluation, which are extracted with Wiktionary and the Interactive Terminology for Europe (IATE) terminology database respectively.

\begin{table}[]\footnotesize
    % \small
    \centering
    \begin{tabular}{c|cc|cc}
    \hline
         & \multicolumn{2}{c|}{\textbf{Wiktionary}} & \multicolumn{2}{c}{\textbf{IATE}} \\
         & BLEU & Accuracy & BLEU & Accuracy \\ \hline
        \baseline & 30.18 & 82.94 & 29.10 & 83.09 \\
        \method & \textbf{30.52} & \textbf{93.67} & \textbf{29.38} & \textbf{95.41} \\
    \hline
    \end{tabular}
    \captionsetup{font={footnotesize}}
    \caption{Lexically constrained machine translation results in English $\rightarrow$ German translation. Accuracy represents the rate of the target phrase appears in the translation output.}
    \label{tab:lexical-constraint}
\end{table}

The results are shown in Table \ref{tab:lexical-constraint}.
\method obtains 10.7\% and 12.3\% absolute lexically constrained translation accuracy improvements over \baseline in Wiktionary and IATE respectively without additional training.
The overall translation performance measured by BLEU scores is improved slightly.
These experiments indicate that \method can incorporate lexical constraint into the translation process effectively.
It is also convenient for \method to incorporate multiple lexical constraints by concatenating multiple bilingual phrases as prompts.

\subsection{Case Study}

Table \ref{tab:case-study} shows a case of a English $\rightarrow$ German translation in the Medical domain.
\method generates different translation outputs with different bilingual phrase prompts.
We can explicitly interfere the translation results by revising prompts.

\subsection{Bonus: Privacy Protection}
\method has one more benefit. Users do not want to leak their specific parallel data for fine-tuning in many cases. Protecting their privacy while improving the translation quality simultaneously needs elaborate design. Users with \method only need to provide a self-built phrase prompts database and can keep the model ignorant with the target corpus.

\section{Conclusion}
\label{sec:conclusion}
% In this work, we identify the difficulties of applying prompt-based learning in machine translation. We propose effective methods to tackle the difficulties.
% Experiments shows that the proposed method improves domain-specific machine translation
% and lexically constrained machine translation without additional training.

In this paper, we propose \method, a zero-shot method to quickly resolve domain adaptation. By automatically building bilingual phrase-level database and retrieving input-relevant prompts by contextual representation, we successfully improve the translation quality in unseen domains. Considering the cost, speed, and even privacy, \method is an important alternative for zero-shot domain adaptation in neural machine translation.

\bibliography{anthology,ref.clean}

\begin{thebibliography}{32}
\expandafter\ifx\csname natexlab\endcsname\relax\def\natexlab#1{#1}\fi

\bibitem[{{Aharoni} and {Goldberg}(2020)}]{aharoni2020unsupervised}
Roee {Aharoni} and Yoav {Goldberg}. 2020.
\newblock Unsupervised domain clusters in pretrained language models.
\newblock In \emph{Proc. of ACL}, pages 7747--7763.

\bibitem[{Bahdanau et~al.(2015)Bahdanau, Cho, and Bengio}]{bahdanau2015neural}
Dzmitry Bahdanau, Kyunghyun Cho, and Yoshua Bengio. 2015.
\newblock Neural machine translation by jointly learning to align and
  translate.
\newblock In \emph{Proc. of ICLR}.

\bibitem[{Bapna and Firat(2019)}]{bapna2019simple}
Ankur Bapna and Orhan Firat. 2019.
\newblock Simple, scalable adaptation for neural machine translation.
\newblock In \emph{Proc. of EMNLP}, pages 1538--1548.

\bibitem[{{Brown} et~al.(2020){Brown}, {Mann}, {Ryder}, {Subbiah}, {Kaplan},
  {Dhariwal}, {Neelakantan}, {Shyam}, {Sastry}, {Askell}, {Agarwal},
  {Herbert-Voss}, {Krueger}, {Henighan}, {Child}, {Ramesh}, {Ziegler}, {Wu},
  {Winter}, {Hesse}, {Chen}, {Sigler}, {Litwin}, {Gray}, {Chess}, {Clark},
  {Berner}, {McCandlish}, {Radford}, {Sutskever}, and
  {Amodei}}]{brown2020language}
Tom~B. {Brown}, Benjamin {Mann}, Nick {Ryder}, Melanie {Subbiah}, Jared
  {Kaplan}, Prafulla {Dhariwal}, Arvind {Neelakantan}, Pranav {Shyam}, Girish
  {Sastry}, Amanda {Askell}, Sandhini {Agarwal}, Ariel {Herbert-Voss}, Gretchen
  {Krueger}, Tom {Henighan}, Rewon {Child}, Aditya {Ramesh}, Daniel~M.
  {Ziegler}, Jeffrey {Wu}, Clemens {Winter}, Christopher {Hesse}, Mark {Chen},
  Eric {Sigler}, Mateusz {Litwin}, Scott {Gray}, Benjamin {Chess}, Jack
  {Clark}, Christopher {Berner}, Sam {McCandlish}, Alec {Radford}, Ilya
  {Sutskever}, and Dario {Amodei}. 2020.
\newblock Language models are few-shot learners.
\newblock In \emph{Proc. of NeurIPS}, volume~33, pages 1877--1901.

\bibitem[{Cai et~al.(2021)Cai, Wang, Li, Lam, and Liu}]{cai2021neural}
Deng Cai, Yan Wang, Huayang Li, Wai Lam, and Lemao Liu. 2021.
\newblock Neural machine translation with monolingual translation memory.
\newblock In \emph{Proc. of ACL}, pages 7307--7318.

\bibitem[{Cao and Xiong(2018)}]{cao2018encoding}
Qian Cao and Deyi Xiong. 2018.
\newblock Encoding gated translation memory into neural machine translation.
\newblock In \emph{Proc. of EMNLP}, pages 3042--3047.

\bibitem[{Chen et~al.(2021)Chen, Chen, Wang, and Li}]{chen2021lexical}
Guanhua Chen, Yun Chen, Yong Wang, and Victor~OK Li. 2021.
\newblock Lexical-constraint-aware neural machine translation via data
  augmentation.
\newblock In \emph{Proc. of IJCAI}, pages 3587--3593.

\bibitem[{Chu and Wang(2018)}]{chu2018survey}
Chenhui Chu and Rui Wang. 2018.
\newblock A survey of domain adaptation for neural machine translation.
\newblock \emph{arXiv preprint arXiv:1806.00258}.

\bibitem[{{Cui} et~al.(2021){Cui}, {Wu}, {Liu}, {Yang}, and
  {Zhang}}]{cui2021template}
Leyang {Cui}, Yu~{Wu}, Jian {Liu}, Sen {Yang}, and Yue {Zhang}. 2021.
\newblock Template-based named entity recognition using bart.
\newblock In \emph{Proc. of ACL}, pages 1835--1845.

\bibitem[{Devlin et~al.(2019)Devlin, Chang, Lee, and
  Toutanova}]{devlin2019bert}
Jacob Devlin, Ming-Wei Chang, Kenton Lee, and Kristina Toutanova. 2019.
\newblock {BERT}: Pre-training of deep bidirectional transformers for language
  understanding.
\newblock In \emph{Proc. of NAACL-HLT}, pages 4171--4186.

\bibitem[{Dinu et~al.(2019)Dinu, Mathur, Federico, and
  Al-Onaizan}]{dinu2019training}
Georgiana Dinu, Prashant Mathur, Marcello Federico, and Yaser Al-Onaizan. 2019.
\newblock Training neural machine translation to apply terminology constraints.
\newblock In \emph{Proc. of ACL}, pages 3063--3068.

\bibitem[{{Dou} and {Neubig}(2021)}]{dou2021word}
Zi-Yi {Dou} and Graham {Neubig}. 2021.
\newblock Word alignment by fine-tuning embeddings on parallel corpora.
\newblock In \emph{Proc. of EACL}, pages 2112--2128.

\bibitem[{Garcia and Firat(2022)}]{garcia2022using}
Xavier Garcia and Orhan Firat. 2022.
\newblock Using natural language prompts for machine translation.
\newblock \emph{arXiv preprint arXiv:2202.11822}.

\bibitem[{He et~al.(2021)He, Huang, Cui, Li, and Liu}]{he2021fast}
Qiuxiang He, Guoping Huang, Qu~Cui, Li~Li, and Lemao Liu. 2021.
\newblock Fast and accurate neural machine translation with translation memory.
\newblock In \emph{Proc. of ACL}, pages 3170--3180.

\bibitem[{Hu et~al.(2019)Hu, Xia, Neubig, and Carbonell}]{hu2019domain}
Junjie Hu, Mengzhou Xia, Graham Neubig, and Jaime~G Carbonell. 2019.
\newblock Domain adaptation of neural machine translation by lexicon induction.
\newblock In \emph{Proc. of ACL}, pages 2989--3001.

\bibitem[{Johnson et~al.(2019)Johnson, Douze, and J{\'e}gou}]{JDH17}
Jeff Johnson, Matthijs Douze, and Herv{\'e} J{\'e}gou. 2019.
\newblock Billion-scale similarity search with gpus.
\newblock \emph{IEEE Transactions on Big Data}, 7(3):535--547.

\bibitem[{{Koehn}(2010)}]{koehn2010statistical}
Philipp {Koehn}. 2010.
\newblock \emph{Statistical Machine Translation}.

\bibitem[{{Koehn} and {Knowles}(2017)}]{koehn2017six}
Philipp {Koehn} and Rebecca {Knowles}. 2017.
\newblock Six challenges for neural machine translation.
\newblock In \emph{Proceedings of the First Workshop on Neural Machine
  Translation}, pages 28--39.

\bibitem[{{Liu} et~al.(2021{\natexlab{a}}){Liu}, {Shen}, {Zhang}, {Dolan},
  {Carin}, and {Chen}}]{liu2021what}
Jiachang {Liu}, Dinghan {Shen}, Yizhe {Zhang}, Bill {Dolan}, Lawrence {Carin},
  and Weizhu {Chen}. 2021{\natexlab{a}}.
\newblock What makes good in-context examples for gpt-3?
\newblock \emph{arXiv preprint arXiv:2101.06804}.

\bibitem[{{Liu} et~al.(2021{\natexlab{b}}){Liu}, {Yuan}, {Fu}, {Jiang},
  {Hayashi}, and {Neubig}}]{liu2021pre}
Pengfei {Liu}, Weizhe {Yuan}, Jinlan {Fu}, Zhengbao {Jiang}, Hiroaki {Hayashi},
  and Graham {Neubig}. 2021{\natexlab{b}}.
\newblock Pre-train, prompt, and predict: A systematic survey of prompting
  methods in natural language processing.
\newblock \emph{arXiv preprint arXiv:2107.13586}.

\bibitem[{Ott et~al.(2019)Ott, Edunov, Baevski, Fan, Gross, Ng, Grangier, and
  Auli}]{ott2019fairseq}
Myle Ott, Sergey Edunov, Alexei Baevski, Angela Fan, Sam Gross, Nathan Ng,
  David Grangier, and Michael Auli. 2019.
\newblock fairseq: A fast, extensible toolkit for sequence modeling.
\newblock In \emph{Proc. of NAACL - Demonstrations}, pages 48--53.

\bibitem[{Peters et~al.(2018)Peters, Neumann, Iyyer, Gardner, Clark, Lee, and
  Zettlemoyer}]{peters2018deep}
Matthew Peters, Mark Neumann, Mohit Iyyer, Matt Gardner, Christopher Clark,
  Kenton Lee, and Luke Zettlemoyer. 2018.
\newblock Deep contextualized word representations.
\newblock In \emph{Proc. of NAACL-HLT}, pages 2227--2237.

\bibitem[{{Schick} et~al.(2020){Schick}, {Schmid}, and
  {Schütze}}]{schick2020automatically}
Timo {Schick}, Helmut {Schmid}, and Hinrich {Schütze}. 2020.
\newblock Automatically identifying words that can serve as labels for few-shot
  text classification.
\newblock In \emph{Proc. of COLING}, pages 5569--5578.

\bibitem[{{Schick} and {Schütze}(2020)}]{schick2020few}
Timo {Schick} and Hinrich {Schütze}. 2020.
\newblock Few-shot text generation with pattern-exploiting training.
\newblock \emph{arXiv preprint arXiv:2012.11926}.

\bibitem[{{Schick} and {Schütze}(2021)}]{schick2021exploiting}
Timo {Schick} and Hinrich {Schütze}. 2021.
\newblock Exploiting cloze-questions for few-shot text classification and
  natural language inference.
\newblock In \emph{Proc. of EACL}, pages 255--269.

\bibitem[{Sennrich et~al.(2016)Sennrich, Haddow, and
  Birch}]{sennrich2016neural}
Rico Sennrich, Barry Haddow, and Alexandra Birch. 2016.
\newblock Neural machine translation of rare words with subword units.
\newblock In \emph{Proc. of ACL}, pages 1715--1725.

\bibitem[{Sun et~al.(2020{\natexlab{a}})Sun, Huang, Dai, and
  Chen}]{sun2020alleviating}
Zewei Sun, Shujian Huang, Xinyu Dai, and Jiajun Chen. 2020{\natexlab{a}}.
\newblock Alleviating the inequality of attention heads for neural machine
  translation.
\newblock \emph{arXiv preprint arXiv:2009.09672}.

\bibitem[{Sun et~al.(2020{\natexlab{b}})Sun, Huang, Wei, Dai, and
  Chen}]{sun2020generating}
Zewei Sun, Shujian Huang, Hao-Ran Wei, Xin-yu Dai, and Jiajun Chen.
  2020{\natexlab{b}}.
\newblock Generating diverse translation by manipulating multi-head attention.
\newblock In \emph{Proc. of AAAI}, volume~34, pages 8976--8983.

\bibitem[{Sun et~al.(2021)Sun, Wang, and Li}]{sun2021multilingual}
Zewei Sun, Mingxuan Wang, and Lei Li. 2021.
\newblock Multilingual translation via grafting pre-trained language models.
\newblock In \emph{Proc. of EMNLP}, pages 2735--2747.

\bibitem[{Sun et~al.(2022)Sun, Wang, Zhou, Zhao, Huang, Chen, and
  Li}]{sun2022rethinking}
Zewei Sun, Mingxuan Wang, Hao Zhou, Chengqi Zhao, Shujian Huang, Jiajun Chen,
  and Lei Li. 2022.
\newblock Rethinking document-level neural machine translation.
\newblock In \emph{Proc. of ACL}, pages 3537--3548.

\bibitem[{{Susanto} et~al.(2020){Susanto}, {Chollampatt}, and
  {Tan}}]{susanto2020lexically}
Raymond~Hendy {Susanto}, Shamil {Chollampatt}, and Liling {Tan}. 2020.
\newblock Lexically constrained neural machine translation with levenshtein
  transformer.
\newblock In \emph{Proc. of ACL}, pages 3536--3543.

\bibitem[{Vaswani et~al.(2017)Vaswani, Shazeer, Parmar, Uszkoreit, Jones,
  Gomez, Kaiser, and Polosukhin}]{vaswani2017attention}
Ashish Vaswani, Noam Shazeer, Niki Parmar, Jakob Uszkoreit, Llion Jones,
  Aidan~N. Gomez, Lukasz Kaiser, and Illia Polosukhin. 2017.
\newblock Attention is all you need.
\newblock In \emph{Proc. of NeurIPS}, pages 5998--6008.

\end{thebibliography}
\bibliographystyle{acl_natbib}

% \appendix

% \section{Example Appendix}
% \label{sec:appendix}

% This is an appendix.

\end{document}